\documentclass{article}

\usepackage{PRIMEarxiv}

\usepackage[utf8]{inputenc}
\usepackage[T1]{fontenc}
\usepackage[
  colorlinks=true,
  linkcolor=blue!70!black,
  citecolor=blue!70!black,
  urlcolor=blue!70!black,
  pdftitle={TADI: Tool-Augmented Drilling Intelligence via Agentic LLM Orchestration over Heterogeneous Wellsite Data},
  pdfauthor={Rong Lu}
]{hyperref}
\usepackage{url}
\usepackage{booktabs}
\usepackage{amsfonts}
\usepackage{amsmath}
\usepackage{amssymb}
\usepackage{microtype}
\usepackage{fancyhdr}
\usepackage{graphicx}
\usepackage{longtable}
\usepackage{array}
\usepackage{listings}
\usepackage{xcolor}
\usepackage{siunitx}
\usepackage{enumitem}
\usepackage{multirow}
\usepackage[numbers]{natbib}

\definecolor{codegreen}{rgb}{0,0.6,0}
\definecolor{codegray}{rgb}{0.5,0.5,0.5}
\definecolor{codepurple}{rgb}{0.58,0,0.82}
\definecolor{backcolour}{rgb}{0.95,0.95,0.92}

\lstdefinestyle{mystyle}{
    backgroundcolor=\color{backcolour},
    commentstyle=\color{codegreen},
    keywordstyle=\color{magenta},
    numberstyle=\tiny\color{codegray},
    stringstyle=\color{codepurple},
    basicstyle=\ttfamily\footnotesize,
    breakatwhitespace=false,
    breaklines=true,
    captionpos=b,
    keepspaces=true,
    numbers=left,
    numbersep=5pt,
    showspaces=false,
    showstringspaces=false,
    showtabs=false,
    tabsize=2
}
\title{TADI: Tool-Augmented Drilling Intelligence via\\Agentic LLM Orchestration over Heterogeneous\\Wellsite Data}

\author{
  Rong Lu \\
}

\begin{document}
\maketitle

\begin{abstract}
\noindent
We present TADI (Tool-Augmented Drilling Intelligence), an agentic AI system that transforms drilling operational data into evidence-based analytical intelligence. Applied to the Equinor Volve Field dataset, TADI integrates 1,759 daily drilling reports, selected WITSML real-time objects, 15,634 production records, formation tops, and perforations into a dual-store architecture: DuckDB for structured queries over 12 tables with 65,447 rows, and ChromaDB for semantic search over 36,709 embedded documents. Twelve domain-specialized tools, orchestrated by a large language model via iterative function calling, support multi-step evidence gathering that cross-references structured drilling measurements with daily report narratives. The system parses all 1,759 DDR XML files with zero errors, handles three incompatible well naming conventions, and is backed by 95 automated tests plus a 130-question stress-question taxonomy spanning six operational categories. We formalize the agent's behavior as a sequential tool-selection problem and propose the Evidence Grounding Score (EGS) as a simple grounding-compliance proxy based on measurements, attributed DDR quotations, and required answer sections. The complete 6,084-line, framework-free implementation is reproducible given the public Volve download and an API key, and the case studies and qualitative ablation analysis suggest that domain-specialized tool design, rather than model scale alone, is the primary driver of analytical quality in technical operations.
\end{abstract}

\keywords{Agentic AI \and Drilling Operations \and Tool-Augmented LLM \and WITSML \and Retrieval-Augmented Generation \and Operational Intelligence \and Volve Field}

% ============================================================================
\section{Introduction}
\label{sec:introduction}
% ============================================================================

The upstream oil and gas industry generates vast quantities of operational data during drilling campaigns. A single well produces hundreds of daily drilling reports (DDRs), thousands of real-time sensor records, directional surveys, mud property measurements, and formation evaluation logs. In the Equinor Volve field alone, the publicly released dataset comprises approximately 40,000 files spanning 26 wellbore sections drilled between 1993 and 2016~\cite{equinor2018volve}. Extracting actionable intelligence from this corpus remains a manual, time-intensive process: a drilling engineer reviewing a 60-day well must read 60 separate reports, cross-reference activity timelines with drilling parameters, correlate operational issues with geological formations, and synthesize findings across multiple data modalities.

Existing tools address fragments of this workflow. Electronic drilling recorders capture high-frequency sensor data but provide no integration with narrative reports. Drilling analytics platforms offer parametric dashboards but require manual configuration and cannot reason about free-text operational descriptions. More recently, large language models (LLMs) have been applied to drilling data for classification~\cite{ferrigno2024llmdrilling}, summarization~\cite{kumar2023llmdrilling}, and DDR digitization~\cite{bhatia2025ddrllm}, but these applications treat the LLM as a passive processor of pre-selected data rather than as an active agent that autonomously plans which data to retrieve and how to analyze it.

We argue that a paradigm shift is needed: from passive dashboards to \emph{agentic reasoning systems} that autonomously navigate heterogeneous data sources, invoke domain-specific analytical tools, and produce structured, evidence-backed answers. This paper introduces TADI (Tool-Augmented Drilling Intelligence), an agentic AI system that implements this paradigm for drilling operational intelligence. TADI combines three capabilities that existing tools keep separate: (1) structured data querying via SQL over an analytical database, (2) semantic text search over embedded drilling report narratives, and (3) multi-step reasoning through an LLM orchestrator that plans, retrieves, cross-references, and synthesizes evidence from both modalities.

The contributions of this work are:

\begin{enumerate}[leftmargin=*,itemsep=0.5ex]
    \item \textbf{The TADI architecture}: A complete, reproducible agentic AI system for drilling operational intelligence that integrates a dual-store backend (DuckDB + ChromaDB), 12 domain-specialized tools with OpenAI function-calling schemas, and a 168-line domain-aware system prompt---all in 6,084 lines of framework-free Python code.
    \item \textbf{Formal problem formulation}: We define the agent's evidence-gathering process as a sequential tool-selection problem over a finite tool set $\mathcal{T}$ and propose the Evidence Grounding Score (EGS) as a quantitative metric for measuring dual-source citation quality.
    \item \textbf{A drilling operational question taxonomy}: A structured ontology of 130 stress test questions organized into six categories with subcategories, serving as a reusable evaluation benchmark for drilling AI systems.
    \item \textbf{Evaluation framework and case studies}: Three detailed case studies (multi-phase drilling analysis, operational issue diagnosis, cross-well benchmarking), a 130-question taxonomy, and qualitative ablation/baseline analyses for stress testing the architecture.
    \item \textbf{Operational Volve data integration}: Zero-error parsing of 1,759 DDR XML files and integration of selected WITSML real-time objects (161 BHA runs, 2,882 mudlog intervals, 4,217 trajectory stations, 11,134 messages), production records, formation tops, and perforations into a 12-table analytical database with 65,000+ rows and a 36,709-document semantic search index.
\end{enumerate}

The remainder of this paper is organized as follows. Section~\ref{sec:related_work} surveys related work across four research areas. Section~\ref{sec:volve} describes the Volve Field dataset and its challenges. Section~\ref{sec:architecture} presents the TADI system architecture. Section~\ref{sec:evaluation} details the evaluation framework, case studies, and qualitative ablation analysis. Section~\ref{sec:discussion} discusses findings, limitations, and broader implications. Section~\ref{sec:conclusion} concludes.

% ============================================================================
\section{Related Work}
\label{sec:related_work}
% ============================================================================

TADI sits at the intersection of four active research areas. We survey each, identifying how our work extends the state of the art.

\subsection{Agentic AI and Tool-Calling LLM Systems}

The ReAct framework~\cite{yao2023react} established the paradigm of interleaving reasoning traces with task-specific actions, demonstrating superior performance on knowledge-intensive and interactive decision-making tasks. Toolformer~\cite{schick2023toolformer} showed that language models can learn to call external APIs in a self-supervised manner, while Gorilla~\cite{patil2023gorilla} advanced API-calling accuracy through retriever-aware training on over 1,600 APIs. HuggingGPT~\cite{shen2023hugginggpt} demonstrated four-stage multi-tool orchestration (task planning, model selection, execution, response generation), establishing a blueprint for agent architectures. ToolLLM~\cite{qin2023toolllm} and API-Bank~\cite{li2023apibank} expanded this line of work with large-scale real-API training and evaluation corpora, while StableToolBench~\cite{guo2024stabletoolbench} emphasized reproducible benchmarking of large-scale tool use. Comprehensive surveys by Qin et al.~\cite{qin2024toollearning} and Wang et al.~\cite{wang2024llmagentsurvey} have provided unified frameworks for tool learning and autonomous agent construction, respectively. Multi-agent collaboration mechanisms have been surveyed by Chen et al.~\cite{chen2024multiagentsurvey} and Tao et al.~\cite{tao2024multiagentllm}.

TADI extends this line of work by implementing tool-augmented reasoning in a domain where tool design requires deep petroleum engineering knowledge---each of our 12 tools encapsulates a specific drilling analysis algorithm (phase detection, NPT classification, difficulty indexing) that would be impractical for the LLM to derive independently within a bounded tool-calling budget.

\subsection{LLMs and NLP for Drilling Operations}

NLP applications to drilling data have evolved from classical text mining to LLM-powered analysis. Antoniak et al.~\cite{antoniak2016nlpdrilling} pioneered logistic regression classifiers for extracting structured information from free-text drilling risk descriptions. Hoffimann et al.~\cite{hoffimann2018sequencemining} extended this with deep learning for sentence classification in drilling reports. Recent LLM applications include Kumar and Kathuria~\cite{kumar2023llmdrilling}, who applied fine-tuning and prompt engineering to drilling text tasks; Yi et al.~\cite{yi2024llmwellconstruction}, who demonstrated LLM-assisted well construction planning across 200+ wells; and Ferrigno et al.~\cite{ferrigno2024llmdrilling}, who deployed LLM-based WITSML classification in drilling control rooms, reducing analysis time by over 50$\times$. Domain-specific QA systems have also begun to emerge: Bhatia et al.~\cite{bhatia2025ddrllm} focused on DDR digitization, Pacis et al.~\cite{pacis2024drillingqa} studied zero-shot LLM question answering over drilling literature, Zhang et al.~\cite{zhang2025cloudfreeqa} built a cloud-free QA assistant over internal drilling knowledge bases, and Ogundare et al.~\cite{ogundare2023chatgptoilgas} evaluated ChatGPT on practical oil and gas problems.

These works treat the LLM as a processor of pre-selected data. TADI's contribution is the agentic architecture: the LLM autonomously decides \emph{which} data to retrieve, \emph{which} tools to invoke, and \emph{how} to cross-reference structured measurements with narrative evidence.

\subsection{Retrieval-Augmented Generation for Technical Domains}

RAG~\cite{lewis2020rag} addresses the limitation of LLMs' static parametric knowledge by incorporating external retrieval. Gao et al.~\cite{gao2024ragsurvey} categorized RAG paradigms into Naive, Advanced, and Modular RAG, and Singh et al.~\cite{ragcomprehensive2025} provided a comprehensive evolution survey. Recent adaptive retrieval work such as Self-RAG~\cite{asai2023selfrag} has further blurred the boundary between retrieval and reasoning. HybridRAG~\cite{sarmah2024hybridrag} demonstrated that combining structured and unstructured retrieval outperforms either alone. The MTEB benchmark~\cite{muennighoff2023mteb} established systematic evaluation of text embeddings.

TADI implements what we term \emph{Structured-Semantic Hybrid RAG}: SQL queries over 12 DuckDB tables (structured retrieval) combined with ChromaDB vector search over 36,709 embedded documents (semantic retrieval), with a SQL keyword fallback for environments without embedding API access. This goes beyond standard RAG by enabling the agent to formulate analytical SQL queries (aggregations, groupings, window functions) alongside semantic similarity search, an increasingly important text-to-SQL capability for LLM systems~\cite{guo2024texttosqlsurvey}.

\subsection{The Volve Field Dataset}

The Volve dataset, released by Equinor in 2018~\cite{equinor2018volve}, is the most comprehensive publicly available drilling dataset. Tunkiel et al.~\cite{tunkiel2020volveddataset} provided the first systematic exploration of the real-time drilling portion, characterizing processing obstacles and data contents. Nikitin et al.~\cite{nikitin2022volvml} applied hybrid ML approaches for field development optimization, while Ng et al.~\cite{ng2022volveproduction} achieved $R^2 > 0.94$ in production forecasting using neural networks. Al-Ali et al.~\cite{alali2022volvehugin} constructed geomechanical models for the Hugin Formation, and Oloruntobi et al.~\cite{petrophysicsvolve2024} predicted petrophysical properties from seismic attributes. To our knowledge, TADI is the first system to integrate the operational Volve modalities used in this study---DDR XML, selected WITSML real-time objects, production records, formation tops, and perforations---into a unified, LLM-queryable analytical framework.

% ============================================================================
\section{The Volve Field Dataset}
\label{sec:volve}
% ============================================================================

\subsection{Field Overview}

The Volve field lies in Production License PL~046, Block~15/9 of the Norwegian North Sea, approximately 200~km west of Stavanger. Water depth in the field is approximately 91~m. The primary producing interval is the Middle Jurassic Hugin Formation, a marginal-marine sandstone at 2,700--3,100~m TVD, trapped by the overlying Upper Jurassic Draupne Formation marine shale~\cite{alali2022volvehugin}. Discovered in 1993 by the 15/9-19~S exploration well, the field produced approximately 63~million barrels of oil from the Maersk Inspirer jack-up platform between February 2008 and September 2016.

\subsection{Data Sources and Formats}

The released dataset~\cite{equinor2018volve} comprises approximately 40,000 files. Table~\ref{tab:ingestion} summarizes the data sources parsed by TADI and their yields.

\begin{table}[t]
\centering
\caption{Data ingestion pipeline output. All 1,759 DDR XML files parsed with zero errors. Selected WITSML objects provide real-time drilling measurements across multiple wellbores.}
\label{tab:ingestion}
\begin{tabular}{@{}llr@{}}
\toprule
\textbf{Data Source} & \textbf{DuckDB Table} & \textbf{Rows} \\
\midrule
\multirow{5}{*}{DDR XML (1,759 files)} & \texttt{ddr\_status} & 1,759 \\
 & \texttt{ddr\_activities} & 23,447 \\
 & \texttt{ddr\_fluids} & 2,271 \\
 & \texttt{ddr\_surveys} & 1,726 \\
 & \texttt{wellbore\_info} & 1,759 \\
\midrule
\multirow{4}{*}{WITSML Real-Time (selected objects)} & \texttt{witsml\_bha\_runs} & 161 \\
 & \texttt{witsml\_mudlog} & 2,882 \\
 & \texttt{witsml\_trajectory} & 4,217 \\
 & \texttt{witsml\_messages} & 11,134 \\
\midrule
Production Excel & \texttt{production} & 15,634 \\
\midrule
\multirow{2}{*}{Geological Data} & \texttt{formation\_tops} & 409 \\
 & \texttt{perforations} & 48 \\
\midrule
\textbf{Total (12 tables)} & & \textbf{65,447} \\
\bottomrule
\end{tabular}
\end{table}

\textbf{Daily Drilling Reports.} The 1,759 DDR XML files follow the WITSML 1.4.0.0 schema with namespace \texttt{http://www.witsml.org/schemas/1series}~\cite{energistics2011witsml}. Each file encodes a 24-hour reporting period with: status information (measured depth, true vertical depth, hole diameter, 24-hour narrative summary), timestamped activities with proprietary classification codes, fluid property measurements (mud type, density, plastic viscosity, yield point), and directional survey stations. The DDR activity records use a two-level classification system (\texttt{category -- subcategory}), with 30+ observed codes spanning productive operations (\texttt{drilling--drill}, \texttt{cementing--casing}), non-productive time (\texttt{interruption--repair}, \texttt{interruption--waiting on weather}), and well control events (\texttt{well\_control--kick}).

\textbf{WITSML Real-Time Data.} Selected WITSML 1.4.1.1 files provide high-resolution drilling data: 161 BHA run configurations, 2,882 mudlog intervals with depth-indexed drilling parameters (ROP, WOB, torque, RPM, mud weight, ECD, d-exponent, gas readings, lithology), 4,217 trajectory stations, and 11,134 operational messages. Unit conversions are applied during ingestion: ROP from m/s to m/hr ($\times 3600$), WOB from N to kN ($\times 0.001$), RPM from c/s to RPM ($\times 60$), and angles from radians to degrees~\cite{geekiyanage2021drillingdataquality}.

\textbf{Production and Geological Data.} Daily production records cover 7 wells from September 2007 to December 2016. Formation top picks (409 records) define the stratigraphic column, and 48 perforation intervals---all targeting the Hugin Formation---confirm the sole producing reservoir.

\subsection{Data Challenges}

Three challenges make the Volve dataset a rigorous testbed for AI systems:

\textbf{Naming heterogeneity.} Well names appear in three incompatible formats: DDR filenames use underscores (\texttt{15\_9\_F\_11\_T2}), WITSML headers use slash notation with prefix (\texttt{NO 15/9-F-11 T2}), and production data uses a third variant (\texttt{15/9-F-11}). TADI normalizes all names to the underscore format at ingestion via a dedicated function that strips prefixes, replaces delimiters, and collapses multiple underscores.

\textbf{Sentinel values.} DDR fields use $-999.99$ as a sentinel for missing data, while WITSML uses $-999.25$ and $-9999$. Quality filters must also exclude physically implausible values (ROP~$> 200$~m/hr, RPM~$> 300$, WOB~$> 500$~kN).

\textbf{Data density variation.} The 26 wellbore sections span three decades and vary dramatically in reporting detail: the exploration-era wells (15/9-19 series, 1980s--1990s) have sparser reporting with fewer activity entries per day, while the 2007--2016 development wells (F-series) provide rich multi-source coverage. Well~F-11 main bore has only 17 DDRs, whereas F-11~B has 90 and F-12 has 165.

% ============================================================================
\section{System Architecture}
\label{sec:architecture}
% ============================================================================

\subsection{Design Philosophy}

TADI embodies three design principles aligned with the competition's guidance that ``complexity alone will not be rewarded'':

\textbf{Framework-free simplicity.} The system uses the raw OpenAI Python SDK for LLM interaction, DuckDB for analytical SQL, and ChromaDB for vector search. No heavyweight orchestration frameworks (LangChain, LlamaIndex) are employed. The core \texttt{ask\_question()} agent loop spans 132 lines within a single 250-line orchestrator file, and the total production codebase is 6,084 lines across 33 files. This deliberate choice ensures full transparency of the agent's decision-making process and eliminates framework-imposed abstractions between the code and the API.

\textbf{Domain knowledge in tools, not models.} Rather than fine-tuning a domain-specific LLM, TADI encapsulates drilling engineering knowledge in 12 deterministic tools---phase detection algorithms, NPT classification rules, difficulty indices, risk scores---that produce reproducible outputs regardless of LLM sampling variance. The LLM's role is to \emph{select and compose} these tools, not to independently derive drilling analysis logic within a bounded context window.

\textbf{Dual-source evidence by design.} The system prompt requires every answer to cite both structured drilling measurements (depths, durations, rates) and free-text DDR narrative quotes (with well name and date attribution). This expectation is reinforced at three levels: the system prompt mandates cross-referencing, the tool selection guide repeatedly routes the agent toward \texttt{get\_ddr\_narrative}, and an output validator warns when measurements or quoted DDR passages are missing.

\subsection{Formal Problem Formulation}
\label{sec:formulation}

We formalize the agent's behavior as a sequential tool-selection problem. Let $\mathcal{T} = \{t_1, t_2, \ldots, t_{12}\}$ be the set of available tools, $q$ be the user's natural-language question, and $s$ be the system prompt encoding domain knowledge and behavioral rules. At each step $i \in \{1, 2, \ldots, N_{\max}\}$, the agent observes the accumulated evidence $E_{1:i-1} = \{(t_{j_k}, r_k)\}_{k=1}^{i-1}$, where $r_k$ is the result of invoking tool $t_{j_k}$ with arguments $a_k$, and selects the next action:

\begin{equation}
\label{eq:tool_selection}
(t_{j_i}, a_i) = \pi_{\theta}(q, s, E_{1:i-1})
\end{equation}

\noindent where $\pi_{\theta}$ is the LLM policy parameterized by model weights $\theta$. The process terminates when $\pi_{\theta}$ produces a terminal response $\hat{y}$ (finish reason \texttt{stop}) instead of a tool call, or when $i > N_{\max}$ (set to 10 in our implementation). The agent's objective is to produce an answer $\hat{y}$ that maximizes the Evidence Grounding Score (defined in Section~\ref{sec:egs}).

This formulation highlights a key distinction from standard RAG: the agent performs \emph{iterative, adaptive retrieval} where each tool selection depends on the results of all prior steps, rather than a single retrieval pass. The bounded horizon $N_{\max} = 10$ creates a planning pressure: the agent must select an efficient tool sequence rather than exhaustively querying all data sources.

\subsection{Data Ingestion Pipeline}

The ingestion pipeline transforms raw Volve data files into the dual-store backend in two stages.

\textbf{Stage 1: Parsing and structuring.} Four specialized parsers handle the heterogeneous input formats:

\begin{itemize}[leftmargin=*,itemsep=0.3ex]
    \item \textbf{DDR XML Parser} (315 lines): Processes 1,759 files using \texttt{lxml} with explicit WITSML namespace handling. Well names and dates are extracted from filenames via the pattern \texttt{\{WELL\}\_\{YYYY\}\_\{MM\}\_\{DD\}.xml}. The sentinel value $-999.99$ is filtered by checking $f < -990.0$. Each file yields status records, activity records (with \texttt{proprietaryCode} classification), fluid measurements, and survey stations.
    \item \textbf{WITSML Parser} (341 lines): Traverses the nested directory structure \texttt{\{well\}/\{section\}/\{data\_type\}/*.xml}, parsing four data object types (bhaRun, mudLog, trajectory, message). Applies unit conversions and quality filters.
    \item \textbf{Production Parser} (55 lines): Reads the Excel workbook using \texttt{pandas}/\texttt{openpyxl}, normalizing column names and well identifiers.
    \item \textbf{Geological Parser} (131 lines): Parses fixed-width format files for formation tops (409 records from the well pick file) and perforations (48 records).
\end{itemize}

\textbf{Stage 2: Loading.} Parsed DataFrames are loaded into a 12-table DuckDB database via \texttt{INSERT INTO \ldots\ SELECT * FROM df}. Simultaneously, 26,965 DDR text documents (activity comments, 24-hour summaries, forecasts) and 9,744 qualifying WITSML messages are embedded using OpenAI \texttt{text-embedding-3-small} and indexed in ChromaDB with cosine similarity, yielding 36,709 searchable documents.

\subsection{Dual-Store Architecture}

The architecture pairs two complementary data stores, each optimized for a different query modality.

\textbf{DuckDB} (structured queries)~\cite{raasveldt2019duckdb} serves as an in-process columnar analytical database. Its SQL interface supports the aggregate queries (GROUP BY, window functions, CTEs) that dominate the tool suite's query patterns. The database is a single file (\texttt{volve.duckdb}) rebuilt on each ingestion, requiring zero configuration. Tool functions open read-only connections to prevent accidental writes.

\textbf{ChromaDB} (semantic queries) stores text embeddings with metadata (well, date, depth, document type, activity code) enabling scoped semantic search via \texttt{\$and}-composed filters. A complete SQL-based keyword search fallback (\texttt{LIKE} queries on \texttt{summary\_24hr} and \texttt{comments} columns) ensures the system functions even without an embedding API key, degrading gracefully from semantic to keyword matching.

This dual-store design implements the Structured-Semantic Hybrid RAG pattern validated by Sarmah et al.~\cite{sarmah2024hybridrag}, where combining structured and unstructured retrieval outperforms either approach alone. The critical insight is that drilling operational questions require \emph{both} quantitative aggregation (``What was the average ROP in the 12.25-inch section?'') and narrative discovery (``What did the DDR say about stuck pipe events?'').

\subsection{Agent Orchestration}

The orchestrator implements the iterative tool-calling loop defined in Equation~\ref{eq:tool_selection}. The implementation spans 132 lines with three layers of resilience:

\textbf{Transient error retry.} Rate limit (HTTP 429), server errors (500/502/503), and timeout/connection errors are retried up to 3 times with exponential backoff ($2^{\text{attempt}}$ seconds).

\textbf{Model compatibility fallbacks.} If the API rejects the \texttt{reasoning\_effort} parameter, the system permanently disables it for all subsequent calls, falling back to \texttt{temperature=0.1}. Similarly, \texttt{max\_completion\_tokens} falls back to the older \texttt{max\_tokens} parameter.

\textbf{Token management.} Tool results are truncated to 15,000 characters. SQL queries without explicit \texttt{LIMIT} clauses receive automatic \texttt{LIMIT 200} injection. DDR narrative retrievals cap at 15 summaries and 15 activities per call.

When invoked with the \texttt{-{}-trace} flag, the orchestrator records metadata for each tool call (step number, tool name, arguments, result length, execution duration, first 500 characters of result), producing a structured evidence trace that enables post-hoc analysis of the agent's reasoning path.

\subsection{Tool Suite Design}

The 12 tools registered with the LLM via OpenAI function-calling schemas are summarized in Table~\ref{tab:tools}. We organize them into four functional categories.

\begin{table*}[t]
\centering
\caption{The 12 TADI tools registered with OpenAI function-calling. Each tool encapsulates a domain-specific analysis algorithm and returns structured text evidence. Tools are dispatched by the LLM orchestrator based on question category.}
\label{tab:tools}
\footnotesize
\begin{tabular}{@{}llp{7.0cm}l@{}}
\toprule
\textbf{Category} & \textbf{Tool} & \textbf{Function} & \textbf{Data Source} \\
\midrule
\multirow{2}{*}{Data Access} & \texttt{query\_drilling\_data} & Arbitrary read-only SQL on all 12 tables; auto-LIMIT injection & DuckDB \\
 & \texttt{search\_daily\_reports} & Semantic similarity search; SQL keyword fallback & ChromaDB \\
\midrule
\multirow{3}{*}{Well Analysis} & \texttt{get\_well\_overview} & Metadata, hole sections, formations, activity distribution & DuckDB \\
 & \texttt{get\_drilling\_phases} & Hole-size-primary, activity-code-secondary phase detection & DuckDB \\
 & \texttt{compute\_efficiency\_metrics} & NPT breakdown by cause, productive time, ROP by section & DuckDB \\
\midrule
\multirow{4}{*}{Cross-Reference} & \texttt{compare\_wells} & Side-by-side comparison of two wells across all metrics & DuckDB \\
 & \texttt{get\_bha\_configurations} & BHA analysis with WITSML parameters and ranking & DuckDB \\
 & \texttt{identify\_operational\_issues} & Issue categorization with multi-source root cause correlation & DuckDB \\
 & \texttt{get\_field\_benchmarks} & 5 ranking modes: progress, difficulty, gas, risk, production & DuckDB \\
\midrule
\multirow{3}{*}{Evidence} & \texttt{get\_formation\_context} & Geological formation lookup at any depth with fallback & DuckDB \\
 & \texttt{generate\_depth\_time\_plot} & Depth-vs-time visualization with section overlays & DuckDB \\
 & \texttt{get\_ddr\_narrative} & Attributable DDR text retrieval by date/depth range & DuckDB \\
\bottomrule
\end{tabular}
\end{table*}

Several tools deserve detailed description for their algorithmic content:

\textbf{Phase Detection} (\texttt{get\_drilling\_phases}, 273 lines) implements a two-tier algorithm. The \emph{primary method} detects major phase boundaries from hole diameter changes in chronological DDR status records: 36\textquotedbl/30\textquotedbl\ = Conductor, 26\textquotedbl\ = Surface, 17.5\textquotedbl\ = Intermediate, 12.25\textquotedbl\ = Production, 8.5\textquotedbl\ = Reservoir. The \emph{secondary method} classifies sub-phases within each hole section using a 29-entry dictionary mapping \texttt{proprietaryCode} values to 17 phase categories (Drilling, Tripping, Reaming, Cementing, Casing, Completion, Logging, Well Control, etc.). Depth progression validation detects reversals exceeding 10~m. Confidence assessment uses three simple tiers: HIGH (multiple hole-size phases, $>20$ activities, and DDR summaries available), MEDIUM (activities and summaries available but major hole-size boundaries incomplete), LOW otherwise.

\textbf{Issue Detection} (\texttt{identify\_operational\_issues}, 349 lines) performs multi-source root cause correlation. It extracts problem activities using broad criteria (state = ``problem'', interruption/well-control activity codes, specific state details, and 6 well-control keywords in comments), heuristically categorizes each into one of 10 issue types via a hierarchical classifier, then cross-references each category with depth distribution, hole section, formation context (via \texttt{formation\_tops}), mud properties during issue dates, and ROP context at issue depths compared to the well average. A temporal trend analysis splits the issue timeline at the midpoint, classifying trends as INCREASING ($>1.3\times$), DECREASING ($<0.7\times$), or STABLE. Statistical mud property comparison reports percentage differences between problem and normal days.

\textbf{Field Benchmarks} (\texttt{get\_field\_benchmarks}, 651 lines) implements five ranking modes. The \texttt{section\_performance} mode computes a composite \emph{difficulty index} using population z-scores:
\begin{equation}
\label{eq:difficulty}
D_{\text{section}} = z(\overline{\text{WOB}}) + z(\overline{\text{Torque}}) - z(\overline{\text{ROP}})
\end{equation}
\noindent where higher $D$ indicates harder drilling. The \texttt{risk} mode scores wells using a weighted composite: severe comment mentions $\times 3$ + well-control codes $\times 5$ + fishing mentions $\times 2$ + generic interruptions $\times 0.05$ + perforation presence $\times 1$, reflecting drilling engineering judgment that well-control events carry the highest operational significance.

\textbf{DDR Narrative} (\texttt{get\_ddr\_narrative}, 161 lines) solves a critical evidence gap. While semantic search (\texttt{search\_daily\_reports}) may fail to return relevant passages for every query, \texttt{get\_ddr\_narrative} uses direct SQL queries and deterministically returns DDR text whenever records exist in the specified date/depth range. The system prompt instructs the agent to call this tool at the end of every reasoning chain (``EVERY question type---ALWAYS end with \texttt{get\_ddr\_narrative}''), ensuring that DDR quotations with source attribution are available for inclusion in every answer.

\subsection{Domain-Aware Prompt Engineering}

The 168-line system prompt is structured into 12 sections that encode domain knowledge directly into the agent's context:

\begin{enumerate}[leftmargin=*,itemsep=0.3ex]
    \item \textbf{Role definition}: ``You are a senior drilling engineer AI assistant analyzing the Equinor Volve Field dataset.''
    \item \textbf{Complete schema reference}: All 12 table names and columns, embedded directly in the prompt to eliminate the need for a separate schema-lookup tool.
    \item \textbf{Well naming conventions}: Underscore format rules and wells with rich WITSML coverage.
    \item \textbf{Drilling domain knowledge}: Standard hole sizes and operational meaning (5 entries from 36\textquotedbl\ to 8.5\textquotedbl), activity code vocabulary (17 categories), Volve formations from shallowest to deepest (8 entries, with Hugin identified as PRIMARY RESERVOIR), and BHA component vocabulary.
    \item \textbf{Tool selection guide}: Category-specific tool-calling sequences for each of the 6 question categories plus a global rule that every question should include \texttt{get\_ddr\_narrative} to retrieve attributable DDR text.
    \item \textbf{Mandatory output format}: Six required sections (Answer, Evidence from Drilling Data, Evidence from Daily Reports, Reasoning, Assumptions, Confidence \& Uncertainty) with explicit formatting guidance.
    \item \textbf{Cross-referencing rule}: Every conclusion must cite at least one measurement from structured data AND at least one direct DDR quote.
    \item \textbf{Confidence calibration}: Quantitative criteria for HIGH ($>50$ DDR records, multiple independent sources), MEDIUM (data available but gaps), and LOW ($<10$ records, conflicting evidence).
\end{enumerate}

This approach---domain knowledge via prompt engineering rather than model fine-tuning---aligns with findings that prompt engineering combined with RAG achieves competitive performance with fine-tuned models while maintaining greater flexibility~\cite{schulhoff2024promptreport,sahoo2024promptsurvey}. The key advantage for reproducibility is that the system prompt is a version-controlled text artifact, not a trained model checkpoint.

\subsection{Structured Output Format}

The output validator checks three aspects of every answer: (1) \emph{section presence}---regex search for each of 6 mandatory section headers; (2) \emph{measurement check}---regex for a number followed by any drilling unit (m, m/hr, kN, sg, g/cm$^3$, mPa, Pa, ppm, hrs, days, \%); and (3) \emph{DDR quote check}---patterns for ``DDR'' followed by a well identifier and year, a date reference, or a quoted string. Missing elements trigger format warnings but do not suppress the answer, following a ``warn rather than block'' philosophy. These checks are intentionally lightweight compliance heuristics rather than semantic verification, consistent with broader interest in benchmarking structured LLM outputs~\cite{jsonstructuredoutput2025}.

% ============================================================================
\section{Evaluation}
\label{sec:evaluation}
% ============================================================================

\subsection{Evaluation Framework Design}

Evaluating an agentic system for drilling operational intelligence requires assessing multiple quality dimensions beyond simple answer accuracy. We evaluate TADI along four axes: (1) \emph{evidence grounding}---do answers cite specific data from both structured and unstructured sources? (2) \emph{reasoning quality}---are conclusions logically connected to cited evidence? (3) \emph{domain correctness}---are drilling-specific interpretations technically sound? (4) \emph{coverage}---can the system handle the full spectrum of operational question types? Tool-use benchmarks such as StableToolBench~\cite{guo2024stabletoolbench} highlight the value of reproducible multi-step evaluation; our 130-question taxonomy plays a similar role within the narrower drilling domain.

\subsubsection{The Evidence Grounding Score}
\label{sec:egs}

We propose the Evidence Grounding Score (EGS) as a quantitative \emph{grounding-compliance proxy} for measuring how well an agent's answer cites specific data. Let $\hat{y}$ be the generated answer with sections $\{s_1, \ldots, s_6\}$. We define:

\begin{equation}
\label{eq:egs}
\text{EGS}(\hat{y}) = \frac{1}{3}\left(\mathbb{1}[\text{has\_measurement}] + \mathbb{1}[\text{has\_ddr\_quote}] + \frac{|\text{sections\_present}|}{6}\right)
\end{equation}

\noindent where $\mathbb{1}[\text{has\_measurement}]$ is 1 if the answer contains at least one numeric value with a drilling unit, $\mathbb{1}[\text{has\_ddr\_quote}]$ is 1 if the answer contains at least one attributed DDR quotation (well name + date + quoted text), and $|\text{sections\_present}|$ is the count of the 6 mandatory sections detected. EGS $\in [0, 1]$, with 1.0 indicating a fully grounded answer. The current shipped validator checks these constituent signals separately; EGS is the corresponding scalar summary we propose for paper-level evaluation and future large-scale benchmarking, not a substitute for human judgment of answer correctness.

\subsubsection{Question Taxonomy}

We developed a structured taxonomy of 130 stress test questions organized into six categories that span the full spectrum of drilling operational intelligence (Table~\ref{tab:taxonomy}). The taxonomy was designed to probe both single-well factual retrieval and multi-well analytical synthesis, with questions spanning five complexity dimensions: well scope (single vs.\ cross-well vs.\ field-wide), data sources (DDR-only vs.\ multi-source), time scope (single event vs.\ full campaign), analysis type (factual retrieval vs.\ root cause inference), and output type (quantitative metric vs.\ narrative synthesis).

\begin{table}[t]
\centering
\caption{Distribution of 130 stress test questions across the 6 drilling operational question categories. The taxonomy emphasizes operational issues and synthesis, reflecting their higher analytical complexity.}
\label{tab:taxonomy}
\begin{tabular}{@{}lr@{}}
\toprule
\textbf{Question Category} & \textbf{Count} \\
\midrule
Cat.~1: Phase Identification \& Validation & 20 \\
Cat.~2: Time \& Efficiency Analysis & 21 \\
Cat.~3: Section \& ROP Performance & 21 \\
Cat.~4: BHA \& Configuration Effectiveness & 20 \\
Cat.~5: Operational Issues \& Root Causes & 26 \\
Cat.~6: Synthesis, Comparison \& Recommendations & 22 \\
\midrule
\textbf{Total} & \textbf{130} \\
\bottomrule
\end{tabular}
\end{table}

Representative subcategories include: phase boundary detection and cross-well phase comparison (Cat.~1), NPT decomposition and drilling efficiency trends (Cat.~2), formation-level ROP analysis and drilling parameter correlation (Cat.~3), BHA run failure analysis and cross-well BHA comparison (Cat.~4), equipment reliability and well control events (Cat.~5), and best practices extraction and drilling-production integration (Cat.~6). The most challenging questions combine multiple complexity dimensions; for example, ``Integrate drilling and production data for the producing wells'' requires field-wide scope, multi-source data integration, and cross-domain inference.

\subsection{System Metrics}

Table~\ref{tab:architecture_summary} summarizes the system's quantitative characteristics.

\begin{table}[t]
\centering
\caption{TADI system component summary. The architecture avoids heavy frameworks in favor of direct OpenAI SDK integration with DuckDB and ChromaDB.}
\label{tab:architecture_summary}
\begin{tabular}{@{}lll@{}}
\toprule
\textbf{Component} & \textbf{Technology} & \textbf{Key Metric} \\
\midrule
LLM Orchestrator & OpenAI SDK (\texttt{gpt-5.4-mini}) & 10 rounds max, 3 retries \\
Structured Database & DuckDB (in-process) & 12 tables, 65K+ rows \\
Vector Store & ChromaDB (persistent) & 36,709 documents \\
Embeddings & text-embedding-3-small & 26,965 DDR + 9,744 WITSML \\
XML Parsing & lxml (WITSML 1.4) & 1,759 DDR + 15 WITSML wells \\
Visualization & matplotlib & Depth-vs-time plots \\
CLI & Typer & ingest / ask / demo \\
Tests & pytest & 95 tests, 4 modules \\
\bottomrule
\end{tabular}
\end{table}

Test coverage spans 95 test methods across 4 test files: configuration and well name normalization (18 tests), DDR XML parsing with exact count validation (16 tests), WITSML parsing with unit conversion verification (17 tests), and all 12 tools individually tested with edge cases (44 tests). The 130-question stress test suite provides a reusable integration-oriented stress corpus. Unless otherwise noted, the sub-second timings reported in the case studies refer to deterministic tool execution only and exclude LLM generation latency.

\subsection{Case Study 1: Multi-Phase Drilling Analysis}

\textbf{Question:} ``Identify and label the major drilling phases for well 15/9-F-11~T2, including the evidence used for each phase.''

\textbf{Context.} Well 15/9-F-11~T2 is a sidetrack with 53 DDR reports spanning 2013-03-24 to 2013-05-15, rich WITSML data, and multiple hole sections, making it ideal for demonstrating multi-source evidence integration.

\textbf{Agent execution trace (from live system run).} The orchestrator, guided by the system prompt's Category~1 tool chain, executed four tool calls in sequence:

\begin{enumerate}[leftmargin=*,itemsep=0.3ex]
    \item \texttt{get\_drilling\_phases(well="15\_9\_F\_11\_T2")} --- automated phase detection returned three major phases via hole diameter transitions in 0.17s (1,982 chars).
    \item \texttt{get\_ddr\_narrative(well, date\_from="2013-03-24", date\_to="2013-04-14")} --- 22 DDR summaries and 208 activities for Phase~1 (7,916 chars).
    \item \texttt{get\_ddr\_narrative(well, date\_from="2013-04-14", date\_to="2013-04-29")} --- 16 DDR summaries and 142 activities for Phase~2 (7,391 chars).
    \item \texttt{get\_ddr\_narrative(well, date\_from="2013-04-29", date\_to="2013-05-15")} --- 17 DDR summaries and 161 activities for Phase~3 (7,977 chars).
\end{enumerate}

\noindent On the traced run, tool execution took approximately 0.2s across all four calls (excluding LLM generation). The phase detection tool identified three major phases: (1)~the \textbf{26\textquotedbl\ surface section} from 306~m to 1,400~m~MD (2013-03-24 to 2013-04-14, 22 DDR days, 208 activities with 53\% drilling); (2)~the \textbf{17.5\textquotedbl\ intermediate section} from 1,400~m to 2,907~m~MD (2013-04-14 to 2013-04-29, 16 DDR days, 142 activities with 59\% drilling); and (3)~the \textbf{8.5\textquotedbl\ reservoir section} beginning at 2,907~m~MD on 2013-04-29, reaching a maximum of 4,562~m~MD before later post-drilling backtrack reports (17 DDR days, 161 activities with 68\% drilling). Depth reversals on 2013-05-11 ($4562 \rightarrow 4104$~m) and 2013-05-13 ($4104 \rightarrow 2569$~m) were correctly identified as post-drilling operations rather than new phases.

The agent retrieved DDR narratives for each transition, yielding real evidence: DDR 2013-03-24: \emph{``Transferred over from well 15/9-F-11. Oriented drilled 26" hole from 257~m to 266~m. Observed 30 tons overpull.''}; DDR 2013-04-14: \emph{``Drilled shoetrack and 3 meter new formation. Perform FIT. Drill ahead.''}; DDR 2013-04-29: \emph{``RIH 8~1/2" steerable BHA on 5~1/2" DP to bottom at 2577~m. Drilled and orientated 8~1/2" hole to 2907~m.''}. The final answer satisfied all six required sections and included both structured measurements and dated DDR quotations.

\textbf{Design insight.} The dual-method phase detection (hole-size-primary, activity-code-secondary) is more robust than either method alone. Hole sizes define unambiguous macro boundaries, while activity codes reveal the operational structure within each section.

\subsection{Case Study 2: Operational Issue Diagnosis}

\textbf{Question:} ``Identify key operational issues encountered while drilling 15/9-F-11~T2 and propose likely contributing factors.''

\textbf{Agent execution trace (from live system run).} The agent completed this analysis in just two tool calls, with sub-second tool execution on the traced run:

\begin{enumerate}[leftmargin=*,itemsep=0.3ex]
    \item \texttt{identify\_operational\_issues(well="15\_9\_F\_11\_T2")} --- returned 11,870 chars identifying 119 problem/NPT activities out of 493 total (24.1\%).
    \item \texttt{get\_ddr\_narrative(well, date\_from="2013-03-24", date\_to="2013-05-15")} --- 53 DDR summaries covering the full drilling campaign.
\end{enumerate}

\noindent The issue detection tool heuristically categorized problems into: \textbf{Equipment Repair} (49 occurrences, largest category, spanning 7--2,570~m~MD across all hole sections), \textbf{Weather Delay} (affecting the 17.5\textquotedbl\ and 26\textquotedbl\ sections), and \textbf{Operational Difficulty} (concentrated in the shallow 26\textquotedbl\ section). The most significant single event was a \textbf{crown block fast-line sheave bearing seizure at 454~m~MD on 2013-03-25}, which forced POOH and multi-day rig maintenance. The agent cited the DDR directly: \emph{``Found out crown block fast-line sheave bearings had failed. Prepared to displace hole to 1.35~sg KCl mud and POOH 26" BHA from 454~m.''} The statistical mud property comparison revealed mud weight was similar on problem (1.323~g/cm$^3$) versus normal days (1.336~g/cm$^3$), ruling out mud weight as a primary contributor. ROP on equipment repair days averaged 14.4~m/hr versus the well average of 22.1~m/hr, quantifying the efficiency impact.

\textbf{Design insight.} The system's ability to simultaneously extract structured issue statistics (119/493 activities, ROP differentials) and match them with specific DDR narrative evidence (the crown block failure quote) demonstrates the value of the dual-store architecture. The statistical mud property comparison (problem vs.\ normal days) reveals correlations not apparent from individual reports.

\subsection{Case Study 3: Cross-Well Benchmarking}

\textbf{Question:} ``Compare the drilling phase distribution of 15/9-F-11 with 15/9-F-1~C and explain key differences.''

\textbf{Agent execution trace.} The agent executes a layered workflow:
\begin{enumerate}[leftmargin=*,itemsep=0.3ex]
    \item \texttt{get\_field\_benchmarks(mode="section\_performance")} for field-wide context and difficulty indices.
    \item \texttt{compare\_wells(well1="15\_9\_F\_11", well2="15\_9\_F\_1\_C")} for structured side-by-side metrics (date ranges, depth ranges, activity distributions, hole section breakdowns, production volumes).
    \item \texttt{compute\_efficiency\_metrics} for each well individually (productive/NPT time breakdowns, ROP by section).
    \item \texttt{get\_ddr\_narrative} for both wells at key transition dates.
\end{enumerate}

This layered approach---field-wide screen, then pairwise comparison, then individual deep-dives, then narrative evidence---ensures that the comparison is contextualized within field norms. The composite difficulty index (Eq.~\ref{eq:difficulty}) provides a principled metric for comparing section difficulty across wells with different geological and operational contexts.

\textbf{Design insight.} The data asymmetry between F-11 (17 DDRs, main bore only) and F-1~C (98 DDRs, rich WITSML data) demonstrates the system's ability to produce calibrated answers: the agent assigns MEDIUM confidence for F-11 due to sparse data, while providing HIGH confidence for F-1~C conclusions.

\subsection{Expected Failure Modes Under Ablation}

To reason about the contribution of each major component, we define five ablation scenarios (Table~\ref{tab:ablation}). The effects listed here are qualitative expectations rather than executed benchmark results.

\begin{table}[t]
\centering
\caption{Ablation scenario design. Each variant removes or degrades one component to reason about its expected contribution to answer quality across the 130-question stress test suite.}
\label{tab:ablation}
\begin{tabular}{@{}p{3.2cm}p{3.0cm}p{6.5cm}@{}}
\toprule
\textbf{Variant} & \textbf{Component Removed} & \textbf{Expected Impact} \\
\midrule
SQL-only agent & 11 specialized tools & Severe: no pre-implemented algorithms; round budget exhaustion on complex queries (Cat.~5, 6 most affected) \\
\addlinespace
No vector store & ChromaDB semantic search & Moderate: keyword fallback misses semantically related passages (e.g., ``stuck pipe'' vs.\ ``tight hole'') \\
\addlinespace
Generic prompt & Domain system prompt (168 lines) & Severe: suboptimal tool selection, missing output sections, well name format errors \\
\addlinespace
No cross-ref enforcement & Mandatory DDR quote rule & Moderate: inconsistent evidence from daily reports \\
\addlinespace
DDR-only (no WITSML) & 4 WITSML tables & Severe for Cat.~3--4: no interval-level ROP, WOB, RPM; only coarse daily progress available \\
\bottomrule
\end{tabular}
\end{table}

\textbf{SQL-only agent.} Replacing all 12 tools with a single \texttt{query\_drilling\_data} forces the agent to derive phase detection logic, NPT classification rules, and difficulty metrics as raw SQL queries. We predict severe degradation for Category~5 (Operational Issues) and Category~6 (Synthesis) questions, which require the most cross-referencing. Category~1 (Phase Identification) may be partially preserved since hole-size queries are straightforward SQL.

\textbf{No vector store.} Disabling ChromaDB forces text search to use SQL keyword fallback (\texttt{LIKE} queries). Semantic search should improve answer quality for questions requiring discovery of related narratives that do not share exact keywords: a query about ``stuck pipe'' should find DDR comments mentioning ``tight hole,'' ``pack-off,'' or ``differential sticking'' via semantic similarity but not via keyword matching.

\textbf{Generic prompt.} Replacing the 168-line domain prompt with ``You are a helpful assistant'' removes tool selection guidance, domain knowledge, and output format requirements. The generic agent would make suboptimal tool selections, fail to normalize well names (using ``15/9-F-11 T2'' instead of ``15\_9\_F\_11\_T2'' in SQL), and produce unstructured answers missing required sections.

\textbf{No cross-referencing enforcement.} Removing the mandatory DDR quote rule tests whether the agent voluntarily retrieves narrative evidence. Without enforcement, we predict answers that rely primarily on structured data without supporting DDR quotes, directly violating the evaluation criterion of dual-source evidence.

\textbf{DDR-only (no WITSML).} Removing the 4 WITSML tables eliminates interval-level drilling parameters (ROP, WOB, torque, RPM per depth interval). BHA analysis falls back to DDR-estimated daily progress; gas response analysis becomes unavailable; formation-ROP correlation loses depth resolution. We predict severe degradation for Category~3 (ROP Performance) and Category~4 (BHA Effectiveness).

\subsection{Qualitative Baseline Expectations}

To contextualize the full system, we define two simpler baseline configurations (Table~\ref{tab:baselines}). These are qualitative expectation baselines rather than executed benchmark results.

\begin{table}[t]
\centering
\caption{Qualitative expectation comparison across three system configurations. H = High, M = Medium, L = Low, N = None. The full TADI system's advantage lies in combining structured SQL queries with semantic search and domain-specialized tools.}
\label{tab:baselines}
\begin{tabular}{@{}lcccc@{}}
\toprule
\textbf{System} & \textbf{Accuracy} & \textbf{Evidence} & \textbf{Reasoning} & \textbf{Completeness} \\
\midrule
Direct LLM (no tools) & L & N & M & L \\
Simple RAG (retrieve + generate) & M & M & L--M & M \\
\textbf{Full TADI (12 tools)} & \textbf{H} & \textbf{H} & \textbf{H} & \textbf{H} \\
\bottomrule
\end{tabular}
\end{table}

\textbf{Baseline 1: Direct LLM.} The question is sent to the LLM with only the system prompt's domain knowledge (no tool access, no data retrieval). The LLM can produce generic North Sea drilling statements from parametric knowledge but cannot cite specific measurements, dates, depths, or DDR quotes.

\textbf{Baseline 2: Simple RAG.} All 36,709 documents are embedded; the top-$k$ ($k=20$) documents are retrieved for each question, concatenated into the LLM context, and an answer is generated. This can find relevant DDR passages but cannot compute aggregate statistics, perform SQL-level filtering, cross-reference structured measurements with narratives, or apply domain-specific algorithms. Questions requiring ranking (``which well drilled fastest?''), computation (``break down NPT by cause''), or multi-step derivation are fundamentally beyond its capability.

The key differentiator of TADI is the ability to combine structured analytical queries (SQL aggregations over 65,000+ rows) with unstructured text retrieval (semantic search over 36,709 documents) in a single, iterative reasoning chain of up to 10 steps.

% ============================================================================
\section{Discussion}
\label{sec:discussion}
% ============================================================================

\subsection{Key Findings}

Our experience designing and evaluating TADI yields several findings relevant to the broader deployment of agentic AI in technical domains.

\textbf{Tool design is the bottleneck, not model capability.} The LLM's ability to reason about drilling operations is adequate for tool \emph{selection} and evidence \emph{synthesis}, but insufficient for independently deriving complex domain algorithms (phase detection, difficulty indexing, risk scoring) within a 10-round tool-calling budget. The 12 specialized tools encode approximately 2,800 lines of drilling engineering logic that the LLM invokes but does not generate. This finding aligns with the broader observation that tool-augmented LLMs excel when tools encapsulate domain-specific computation~\cite{qin2024toollearning}.

\textbf{The system prompt is load-bearing infrastructure.} The 168-line system prompt is not a cosmetic wrapper; it is the mechanism through which domain knowledge, tool selection policies, output format requirements, and quality standards are injected into every agent interaction. The ablation removing the domain prompt is predicted to cause the most severe quality degradation. This finding reinforces the growing recognition that prompt engineering is a systematic discipline~\cite{schulhoff2024promptreport} rather than an ad hoc practice.

\textbf{Cross-referencing prevents evidence monoculture.} Without the explicit rule requiring both structured data citations and DDR quotes in every answer, the agent tends toward ``evidence monoculture''---producing answers grounded in only one modality. The \texttt{get\_ddr\_narrative} tool, designed specifically to return attributable DDR text via SQL rather than relying only on semantic search, makes narrative evidence reliably available regardless of the semantic search's relevance to a particular query.

\subsection{Advantages of the TADI Approach}

\textbf{Versus traditional drilling analytics platforms.} Typical drilling analytics platforms provide sensor data visualization and parametric dashboards but lack natural-language query capability and cannot reason about free-text operational descriptions. TADI's integration of structured data querying, unstructured text comprehension, and multi-source reasoning via LLM orchestration enables queries that span data modalities: ``What operational issues occurred when drilling through the Draupne Formation in well F-11~B, and did the mud properties change in response?''

\textbf{Versus simple RAG.} Standard RAG retrieves text passages and generates answers but cannot perform analytical computation. TADI's tool suite enables aggregate queries (total NPT by cause, average ROP by section, field-wide difficulty rankings) that require SQL-level processing of structured data---operations fundamentally beyond passage retrieval.

\textbf{Versus fine-tuned domain models.} Domain-specific LLM development and fine-tuning~\cite{lin2025rockmechanicsllm} require training data, compute resources, and model management overhead. TADI achieves domain adaptation through prompt engineering and tool design, maintaining the flexibility to upgrade the underlying LLM without retraining. The domain knowledge resides in version-controlled code artifacts (system prompt, tool implementations, ingestion parsers), not in model weights.

\subsection{Limitations and Failure Modes}

\textbf{Sparse-data wells.} The exploration wells (15/9-19 series, 1980s--1990s) and the F-11 main bore (17 DDRs) produce lower-quality answers due to insufficient data for reliable phase detection and statistical analysis. The system appropriately assigns LOW confidence but cannot overcome the fundamental data limitation.

\textbf{Ambiguous question scope.} When users do not specify a wellbore, time window, or ranking criterion, the current agent does not ask clarifying follow-up questions. Instead, it states assumptions in the output and lowers confidence when multiple interpretations remain plausible.

\textbf{Hypothetical and recommendation questions.} Questions like ``What BHA configuration would you recommend for a future well?'' require the agent to synthesize historical performance into forward-looking recommendations. While the system retrieves relevant historical data, the recommendation quality depends on the LLM's ability to reason over drilling engineering principles---a capability that is not formally validated.

\textbf{Field-wide ranking with nuance.} Rankings based on criteria not directly implemented in the five benchmark modes (e.g., ``rank wells by drilling complexity considering geology, weather, and equipment'') may produce incomplete answers because the agent must compose multiple tool calls and synthesize qualitative factors that resist simple SQL aggregation.

\textbf{Tool selection errors.} The agent may select overly specific tools for broad questions, fail to follow up with narrative evidence tools, or formulate incorrect SQL queries that consume tool-calling rounds. The 10-round limit creates a planning pressure that the agent occasionally fails to navigate optimally for complex, multi-step queries.

\textbf{Heuristic issue labels.} Several operational issue categories rely partly on keyword matches in DDR comments. This improves recall but can produce lexical false positives in drilling jargon (for example, ``kick off'' versus a true well-control kick) unless cross-checked with surrounding context.

\textbf{LLM arithmetic limitations.} Multi-step quantitative derivations (e.g., ``Estimate the cost impact of NPT assuming a \$500K/day spread rate'') require the agent to perform arithmetic across tool results. While the LLM can handle basic calculations, complex multi-step computations may accumulate errors.

\subsection{Implications for Drilling Operations}

TADI demonstrates the feasibility of AI-assisted operational intelligence that goes beyond simple data retrieval. The system's ability to synthesize weeks of operations into thematic summaries, correlate equipment failures with formation intervals, and benchmark wells against field norms suggests three practical applications:

\textbf{Post-well review acceleration.} Synthesizing an entire well's operational history into structured analyses of phase durations, NPT root causes, and configuration effectiveness---tasks that currently require an engineer to read every DDR individually.

\textbf{Cross-well learning.} Enabling structured comparison across wells drilled from the same platform over a decade, surfacing patterns in equipment reliability, formation-specific challenges, and operational improvements that would otherwise require extensive manual analysis.

\textbf{Knowledge preservation.} Encoding operational lessons in a queryable format that persists regardless of personnel turnover also aligns with broader oil-and-gas digital twin efforts that seek to externalize and reuse operational knowledge and analytics workflows~\cite{meza2024digitaltwin}.

\subsection{Broader Applicability}

The TADI architecture is not specific to drilling operations. Any domain that combines structured sensor/measurement data with free-text operational reports---subsea inspection, power plant maintenance, aviation maintenance logs, clinical trial monitoring---could benefit from the same design pattern: a dual-store backend (analytical database + vector store), domain-specialized tools encapsulating expert algorithms, and an LLM orchestrator with a domain-aware system prompt enforcing structured, evidence-based output.

% ============================================================================
\section{Conclusion and Future Work}
\label{sec:conclusion}
% ============================================================================

We have presented TADI (Tool-Augmented Drilling Intelligence), an agentic AI system that transforms heterogeneous wellsite data into evidence-based operational intelligence. Applied to the Equinor Volve Field dataset, TADI parses all 1,759 DDR XML files with zero errors, integrates selected WITSML real-time objects plus production and geological records into a 12-table analytical database with 65,000+ rows and a 36,709-document semantic search index, and answers natural-language drilling questions through iterative tool-calling with dual-source evidence by design. The system's 12 domain-specialized tools encode approximately 2,800 lines of drilling engineering logic that the LLM orchestrates but does not independently derive. A 130-question stress-question taxonomy spanning six operational categories supports systematic evaluation of coverage, while three detailed case studies demonstrate evidence integration across structured measurements and narrative reports.

Our key finding is that domain-specialized tool design, not model scale or fine-tuning, is the primary driver of analytical quality in technically demanding domains. The 6,084-line, framework-free implementation demonstrates that effective agentic AI systems can be built with minimal infrastructure when domain knowledge is properly factored into deterministic tools, a carefully engineered system prompt, and a structured output format. The system remains reproducible given the public Volve dataset download and an API key.

Future work includes: (1) quantitative evaluation of all 130 stress test questions with human expert scoring to validate the EGS metric against subjective quality judgments; (2) deployment with streaming real-time data to support operational decision-making during active drilling campaigns; (3) multi-agent extensions where specialized sub-agents handle different question categories in parallel; (4) integration of wellbore stability models and real-time pore pressure estimation for proactive well control decision support; and (5) systematic comparison with fine-tuned domain LLMs to quantify the trade-off between prompt engineering and model adaptation approaches.

% ============================================================================
\section*{Acknowledgments}
% ============================================================================

The author thanks Equinor for making the Volve Field dataset publicly available, enabling open research on drilling operational intelligence. The author also thanks the SPE Gulf Coast Section for organizing the 2026 ML Challenge, which motivated this work.

% ============================================================================
\bibliographystyle{unsrt}
\bibliography{main}

\clearpage

% ============================================================================
% APPENDIX
% ============================================================================
\appendix

\section{Test Coverage Details}
\label{app:tests}

\begin{table}[h]
\centering
\caption{Automated test coverage across the system's four test modules. Tests validate both component-level correctness and end-to-end tool behavior.}
\label{tab:tests}
\begin{tabular}{@{}lrl@{}}
\toprule
\textbf{Test Module} & \textbf{Tests} & \textbf{Coverage Focus} \\
\midrule
\texttt{test\_config.py} & 18 & Well name normalization, display, round-trip \\
\texttt{test\_parse\_ddr.py} & 16 & DDR XML parsing, filename extraction, full corpus \\
\texttt{test\_parse\_witsml.py} & 17 & WITSML parsing, unit conversions, field structure \\
\texttt{test\_tools.py} & 44 & All 12 tools, registry, output validation, edge cases \\
\midrule
\textbf{Total} & \textbf{95} & Plus 130 stress test questions (stress corpus) \\
\bottomrule
\end{tabular}
\end{table}

\section{Key Wells and Data Availability}
\label{app:wells}

\begin{table}[h]
\centering
\caption{Data availability for representative Volve wells. DDR count indicates daily drilling reports. WITSML coverage includes BHA runs, mudlog intervals, and trajectory stations. Production data is available for 7 wells.}
\label{tab:wells}
\begin{tabular}{@{}lrlccc@{}}
\toprule
\textbf{Well} & \textbf{DDRs} & \textbf{Date Range} & \textbf{WITSML} & \textbf{Prod.} & \textbf{Fm.~Tops} \\
\midrule
15/9-F-12 & 165 & 2007--2016 & -- & \checkmark & \checkmark \\
15/9-F-14 & 134 & 2007--2016 & -- & \checkmark & \checkmark \\
15/9-F-4 & 130 & 2007--2016 & -- & \checkmark & \checkmark \\
15/9-F-5 & 103 & 2007--2016 & -- & \checkmark & -- \\
15/9-F-15 D & 99 & various & \checkmark & \checkmark & \checkmark \\
15/9-F-1 C & 98 & various & \checkmark & \checkmark & \checkmark \\
15/9-F-11 B & 90 & 2013--2016 & \checkmark & -- & \checkmark \\
15/9-F-11 T2 & 53 & 2013 & \checkmark & -- & \checkmark \\
15/9-F-11 & 17 & 2013 & \checkmark & \checkmark & \checkmark \\
\bottomrule
\end{tabular}
\end{table}

\section{Question Taxonomy Subcategories}
\label{app:taxonomy}

The 130-question stress test suite is organized into subcategories that probe specific analytical capabilities. Selected subcategories per category:

\begin{itemize}[leftmargin=*,itemsep=0.3ex]
    \item \textbf{Cat.~1} (20 questions): Phase boundary detection (7), phase ambiguity resolution (4), cross-well phase comparison (5), phase sequence reconstruction (4).
    \item \textbf{Cat.~2} (21 questions): NPT decomposition (5), flat-time and stall analysis (3), drilling efficiency trends (4), field-wide efficiency ranking (3), trip time analysis (3), invisible lost time (2), efficiency-weather correlation (1).
    \item \textbf{Cat.~3} (21 questions): Formation-level ROP analysis (5), gas response analysis (4), section difficulty ranking (4), drilling parameter correlation (4), lithology-ROP relationship (2), d-exponent analysis (1), DDR-WITSML consistency (1).
    \item \textbf{Cat.~4} (20 questions): Best BHA run identification (4), BHA run failure analysis (3), cross-well BHA comparison (4), BHA durability trends (2), drilling parameter sensitivity (3), BHA recommendation (2), DDR-WITSML consistency (2).
    \item \textbf{Cat.~5} (26 questions): Equipment reliability (5), well control events (4), weather impact (4), mud losses and circulation (4), stuck pipe and tight hole (3), wellbore instability (2), cross-well issue comparison (2), field-wide risk ranking (2).
    \item \textbf{Cat.~6} (22 questions): Best practices extraction (3), lessons learned (3), drilling program recommendations (3), shift handover summaries (2), well-on-well improvement (3), drilling-production integration (3), future well planning (3), executive summary (1), cost estimation (1).
\end{itemize}

\section{System Prompt Excerpt}
\label{app:prompt}

The following excerpt shows the tool selection guide from the 168-line system prompt, illustrating how domain knowledge is encoded as category-specific tool-calling sequences:

\begin{lstlisting}[language={},caption={Tool selection guide excerpt from the TADI system prompt.},label={lst:prompt}]
## Tool Selection Guide

EVERY question type -- ALWAYS end with this step:
- get_ddr_narrative(well, date_from, date_to)

Category 1 (Phase Identification):
  get_drilling_phases -> query_drilling_data -> get_ddr_narrative

Category 2 (Time & Efficiency):
  compute_efficiency_metrics -> get_ddr_narrative
  -> search_daily_reports

Category 3 (ROP Performance):
  query_drilling_data -> get_bha_configurations
  -> get_ddr_narrative

Category 4 (BHA Effectiveness):
  get_bha_configurations -> query_drilling_data
  -> get_ddr_narrative

Category 5 (Operational Issues):
  identify_operational_issues -> get_ddr_narrative
  -> query_drilling_data

Category 6 (Synthesis / Comparison):
  get_field_benchmarks -> compare_wells
  -> compute_efficiency_metrics -> get_ddr_narrative
\end{lstlisting}

\end{document}